\documentclass[runningheads]{llncs}
\usepackage{epsfig}
\usepackage{graphicx}
\usepackage{epstopdf}
\usepackage{amsmath,amssymb} % define this before the line numbering.
\usepackage{color}
\usepackage{subfigure}
\usepackage{enumitem}
\usepackage[width=122mm,left=12mm,paperwidth=146mm,height=193mm,top=12mm,paperheight=217mm]{geometry}

\begin{document}
% \linenumbers
\pagestyle{headings}
\mainmatter

\title{MS-Celeb-1M: 
A  Dataset and Benchmark for 
Large-Scale Face Recognition
%Identification
%Recognizing One Million Celebrities
%in the Real World
%in the Wild
} % Replace with your title

\titlerunning{MS-Celeb-1M: 
A  Dataset and Benchmark for 
Large-Scale Face Recognition}

%\titlerunning{ECCV-16 submission ID \ECCV16SubNumber}

%\authorrunning{ECCV-16 submission ID \ECCV16SubNumber}

\author{Yandong Guo, Lei Zhang, Yuxiao Hu, Xiaodong He, Jianfeng Gao}

\institute{Microsoft Research\\%,\\
%	Microsoft\\
	\email{ \{yandong.guo,leizhang,yuxiao.hu,xiaohe,jfgao\}@microsoft.com}
}
%\titlerunning{title running}

%\authorrunning{authors running}

%\author{Authors}
%\institute{Institute}
{\addtocounter{footnote}{-1}\let\thefootnote\relax\footnotetext{This paper is published at ECCV 2016.}}
%\blfootnote{This paper is published at ECCV 2016.}
\maketitle

\begin{abstract}
In this paper, we design a benchmark task and provide the associated datasets 
for recognizing face images
and link them to corresponding entity keys in a knowledge base. 
More specifically, 
we propose a benchmark task to recognize one million celebrities from their face images, 
by using all the possibly collected face images of this individual on the web as training data. 
The rich information provided by the knowledge base 
helps to conduct disambiguation and improve the recognition accuracy, 
and contributes to various real-world applications,
such as image captioning and news video analysis. 
Associated with this task, 
we design and provide concrete measurement set, evaluation protocol, as well as training data. 
We also present in details our experiment setup and report promising baseline results. 
Our benchmark task could lead to one of the largest classification problems in computer vision.
To the best of our knowledge, our training dataset, which contains $10$M images in version $1$,
is the largest publicly available one in the world.

\end{abstract}

\section{Introduction}
In this paper, 
we design a benchmark task as to recognize one million celebrities from their face images
and identify them by linking to the unique entity keys in a knowledge base. 
We also construct associated datasets to train and test for this benchmark task.  
Our paper is mainly to close the following two gaps in current face recognition, as reported in \cite{Yandong:Celeb}. 
First, there has not been enough effort in determining the identity of a person from a face image
with disambiguation, especially at the web scale.
The current face identification task mainly focuses on finding similar images (in terms of certain types of distance metric)
for the input image, 
rather than answering questions such as 
``who is in the image?''
and ``if it is Anne in the image, which Anne?''. 
This lacks an important step of ``recognizing''. 
%\yandongc{maybe we can structural in a different way, why celebrity? why one million? why freebase? }
The second gap is about the scale.  
%and measurement 
The
publicly available datasets
are much smaller 
than that being used privately in industry, such as Facebook
%\cite{FaceBook_2014}\cite{FaceBook_2015} 
\cite{FaceBook_2014,FaceBook_2015}
and Google \cite{Google_Face},
as summarized in Table \ref{table:datasets}. 
Though the research in face recognition highly desires large datasets
consisting of many distinct people, 
such large dataset is not easily or publicly accessible to most researchers. 
This greatly limits the contributions from research groups, especially in academia. 
%Though MegaFace in \cite{UW_megaface} blends one million common images into the gallery set, 
%the identification task is still evaluated on a relatively small group of people. 

\begin{figure}
\centering
    \includegraphics[width=0.66\columnwidth]{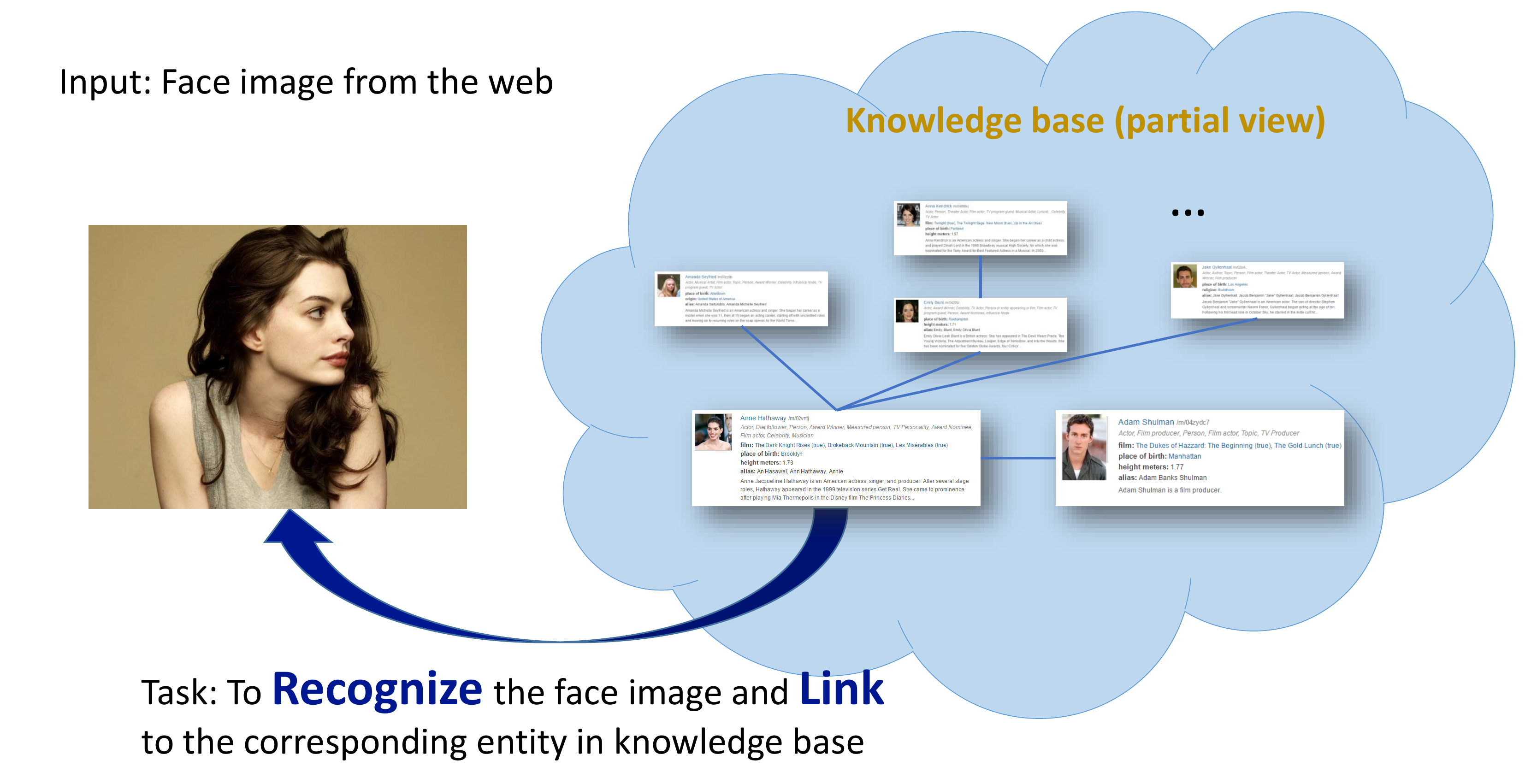}
  \caption{An example of our face recognition task. 
Our task is 
to recognize the face in the image
and then link this face with the corresponding
%person's 
entity key in the knowledge base.
By recognizing the left image to be ``Anne Hathaway'' and linking to the entity key, 
we know she is an American actress born in 1982, who has played Mia Thermopolis in The Princess Diaries, 
not the other Anne Hathaway who was the wife of William Shakespeare. 
Input image is from the web. \protect\footnotemark }
\label{Figure:flExample}
\end{figure}

Our benchmark task has the following properties. 
First, we define our face recognition as 
to determine the identity of a person from his/her face images. 
More specifically, 
we introduce a {\bf{knowledge base}} into face recognition,
since 
the recent advance in knowledge bases has demonstrated incredible capability of 
providing accurate identifiers and rich properties for celebrities. 
Examples include Satori knowledge graph in Microsoft and ``freebase'' in \cite{freebase}. 
Our face recognition task is demonstrated in Fig. \ref{Figure:flExample}. 

\addtocounter{footnote}{-1}
\footnotetext{Image resource: http://www.hdwallpapers.in/anne\_hathaway\_2-wallpapers.html, retrieved by image.bing.com.}

Linking the image with an entity key in the knowledge base,
rather than an isolated string for a person's name
naturally solves the disambiguation issue in the traditional face recognition task. 
Moreover, the linked entity key is associated with rich and comprehensive property information in the knowledge base, 
which makes our task more similar to human behavior compared with traditional face identification,
since retrieving the individual's name as well as the associated information naturally takes place when humans are viewing a face image. 
The rich information makes 
our face recognition task practical and beneficial to many real applications, 
including image search, ranking, caption generation, image deep understanding, etc. 
%Our task has the potential to complete the last step of ``recognizing'' a face image.

Second, our benchmark task targets at recognizing {\bf{celebrities}}. 
Recognizing celebrities, rather than a pre-selected private group of people, 
represents public interest and could be directly applied to a wide range of real scenarios. 
Moreover, 
only  
with popular celebrities,
we can leverage the existing information (e.g. name, profession) in the knowledge base and the information on the web
to build a large-scale dataset which is publicly available for training, measurement, and re-distributing under certain licenses. 
The security department may have many labeled face images for criminal identification, 
but the data can not be publicly shared. 

Third, we select {\bf{one million}} celebrities from freebase and provide their associated entity keys, 
and encourage researchers to build recognizers to
%robustly 
identify each people entity. 
Considering each entity as one class may lead to, to the best of our knowledge, 
the largest classification problem in computer vision. 
The clear definition and mutually exclusiveness of these classes 
%are ensured by 
are supported by
the unique entity keys and their associated properties provided by the knowledge base, 
since in our dataset, 
there are a significant amount of celebrities having same/similar names. 
This is different from 
generic image classification,
where to obtain a large number of exclusive classes with clear definition itself is a challenging and open problem \cite{ILSVRC15}. 

The large scale of our problem naturally introduces the following attractive challenges. 
With the increased number of classes, the inter-class variance tends to decrease. 
There are celebrities look very similar to each other (or even twins) in our one-million list. 
Moreover, large intra-class variance is introduced by 
popular celebrities with millions of images available, 
as well as celebrities with very large appearance variation (e.g., due to age, makeups, or even sex reassignment surgery). 

In order to evaluate the performance of our benchmark task, we provide concrete measurement set and evaluation protocol. 
Our measurement set 
consists of images for a subset of celebrities in our one-million celebrity list. 
The celebrities are selected in a way that,
our measurement set mainly focuses on popular celebrities to represent
the interest of real application and users, 
while the measurement set still maintains enough (about $25\%$) tail celebrities
to encourage the performance on celebrity coverage. 
We manually label images for these celebrities carefully. 
The correctness of our labeling is ensured by deep research on the web content, consensus verification, 
and multiple iterations of carefully review.  
\iffalse
Our measurement set has two subset (covers the same celebrity subset). 
One is focused on evaluating the generalization ability of the model 
on complex situations such as pose/age/resolution/special effects/etc, 
while the other one is focused on evaluating the celebrity coverage. \
More details is introduced in the following sections. 
\fi
In order to %avoid manually labeling onto our measurement set, 
make our measurement more challenging, 
we blend a set of distractor images
with this set of carefully labeled images. 
The distractor images 
are images of other celebrities or ordinary people on the web, 
which are mainly used to hide the celebrities we select in the measurement. 

Along with this 
challenging yet attractive large scale benchmark task proposed, 
we also provide a very large training dataset %sufficiently large
to facilitate the 
task. 
The training dataset contains about $10$M images
for $100$K top celebrities selected from our one-million celebrity list in terms of their web appearance frequency. 
Our training data is, to the best of our knowledge, the largest publicly available one in the world, as shown in Table \ref{table:datasets}. 
We plan to further extend the size in the near future. 
For each of the image in our training data, 
we provide the thumbnail of the original image and cropped face region from the original image (with/without alignment). 
%and aligned face region from the original image. 
This is to maximize the convenience for the researchers to investigate using this data. 

With this training data, we trained 
a convolutional deep neural network with the classification setup (by considering each entity as one class). 
The experimental results show that without extra effort in fine-tuning the model structure, we recognize $44.2\%$ of the images in the measurement set with the precision $95\%$ (hard case, details provided in section \ref{sec:CelebRec}). 
We provide the details of our experiment setup and experimental results to serve as a very promising baseline in section \ref{sec:CelebRec}. 

\subsubsection{Contribution Summary}
%In summary, 
Our contribution in this paper is summarized as follows. 
\begin{itemize}
\item We design a benchmark task: to recognize one million celebrities from their face images, 
and link to their corresponding entity keys in freebase \cite{freebase}.
\item We provide the following datasets,\protect\footnotemark
\footnotetext{Instructions and download links: http://msceleb.org}
\begin{itemize}[label={--}]
\item One million celebrities selected from freebase with corresponding entity keys
%(MID)
, and a snapshot for freebase data dumps;
\item Manually labeled measurement set 
with carefully designed evaluation protocol;
\item A large scale training dataset, with face region cropped and aligned (to the best of our knowledge, the largest publicly available one). 
\end{itemize}
\item We provide promising baseline performance with our training data to inspire more research effort on this task.   
\end{itemize}

Our benchmark task could lead to a very large scale classification problem in computer vision with meaningful real applications. 
This benefits people in experimenting different recognition models (especially fine-grained neural network)
with the given training/testing data. 
Moreover, we encourage people to bring in more outside data and evaluate experimental results in a separate track. 

\section{Related works}

Typically, there are two types of tasks for face recognition. 
One is very well-studied, called face verification, 
which is to determine whether two given face images belong to the same person.
Face verification has been heavily investigated. 
One of the most widely used measurement sets for verification is Labeled Faces in the Wild (LFW)
%in \cite{LFWTech} \cite{LFWTechUpdate},
in \cite{LFWTech,LFWTechUpdate},
which provides $3000$ matched face image pairs and $3000$ mismatched face image pairs,
and allows researchers to report verification accuracy with different settings.
%After years of actively report, %we observe that 
% (supervised, unsupervised, outside data, etc..). 
%for this verification measurement set, 
The best performance on LFW datasets has been frequently updated in the past several years. 
Especially, with the ``unrestricted, labeled outside data'' setting, 
multiple research groups have claimed higher accuracy than human performance
for verification task on LFW
%\cite{Google_Face}\cite{Xiaoou_Deep3}. 
\cite{Google_Face,Xiaoou_Deep3}. 

Recently, 
the interest in the other type of face recognition task, face identification, has greatly increased
\cite{Xiaoou_Deep3,FacePP_ACM,UW_MegaFace,FaceBook_2015}. 
%\cite{Xiaoou_Deep3},\cite{FacePP_ACM},\cite{UW_MegaFace}, \cite{FaceBook_2015}. 
For typical face identification problems, 
two sets of face images are given, called gallery set and query set. 
Then the task is, for a given face image in the query set, to find the most similar faces in the gallery image set. 
When the gallery image set only has a very limited number (say, less than five) of face images for each individual, the most effective solution is still to learn a generic feature which can tell whether or not two face images are the same person, which is essentially still the problem of face verification.
%Efforts include \cite{FaceIdentification}, \cite{FacePP_ACM} and \cite{UW_MegaFace}
Currently, the MegaFace in \cite{UW_MegaFace} might be one of the most difficult face identification benchmarks. 
The difficulty of MegaFace mainly comes from the up-to one million distractors blended in the gallery image set. 
Note that 
the query set in MegaFace
are selected 
from images from FaceScrub \cite{FaceScrub} and FG-NET \cite{FGNet}, which contains 
$530$ and $82$ persons respectively. 

Several 
datasets have been published to facilitate the training  
for the face verification and identification tasks. 
Examples 
include LFW \cite{LFWTech,LFWTechUpdate}, Youtube Face Database (YFD) \cite{Youtube}, 
CelebFaces+ \cite{Xiaoou_Deep1},
and CASIA-WebFace \cite{CASIA_WebFace}. 
In LFW, $13000$ images of faces were collected from the web, and then carefully labeled with celebrities' names. 
The YFD contains $3425$ videos of $1595$ different people. 
The CelebFace+ dataset contains $202, 599$ face images of $10, 177$ celebrities. 
People in CelebFaces+ and LFW are claimed to be mutually exclusive.
The CASIA-WebFace \cite{CASIA_WebFace} is currently the largest dataset which is publicly available,
with about $10$K celebrities, and $500$K images. 
A quick summary is listed in Table \ref{table:datasets}. 
\begin{table}
\caption{Face recognition datasets}
\label{table:datasets}
%\centering
\begin{center} 
%% Some packages, such as MDW tools, offer better commands for making tables
%% than the plain LaTeX2e tabular which is used here.
\begin{tabular}{|c||c|c|c|}
\hline
Dataset & Available &  people &  images \\
\hline
%\bf{Ours} & public & $100$K & about $10000$ K \\
IJB-A \cite{2015_CVPR_FaceData} & public & $500$ & $5712$  \\
\hline
LFW \cite{LFWTech,LFWTechUpdate} & public & $5$K & $13$K \\
\hline
YFD \cite{Youtube}& public & 1595 & 3425 videos \\
\hline
CelebFaces \cite{Xiaoou_Deep1} & public & $10$K & $202$K  \\
\hline
CASIA-WebFace \cite{CASIA_WebFace}& public & $10$K & $500$K  \\
\hline
%\bf{Ours} & public & $100$K & about $10000$ K \\
\bf{Ours} & \bf{public} & \bf{$100$K} & about \bf{$10$M}  \\
\hline
Facebook & private & $4$K & $4400$K  \\
\hline
%Google & private & $8000$ K & $100-200$ M \\
Google & private & $8$M & $100$-$200$M \\
\hline
\end{tabular}
\end{center}
\end{table}
\iffalse
\caption{Table 1. Face recognition datasets}
\label{table:datasets}
%\centering
\begin{center} 
%% Some packages, such as MDW tools, offer better commands for making tables
%% than the plain LaTeX2e tabular which is used here.
\begin{tabular}{|c||c|c|c|c|}
\hline
Dataset & Available &  people &  images & Detection Independent\\
\hline
%\bf{Ours} & public & $100$K & about $10000$ K \\
IJB-A \cite{2015_CVPR_FaceData} & public & $500$ & $5712$ & Yes \\
\hline
LFW \cite{LFWTech,LFWTechUpdate} & public & $5$K & $13$K & No\\
\hline
YFD \cite{Youtube}& public & 1595 & 3425 videos & No\\
\hline
CelebFaces \cite{Xiaoou_Deep1} & public & $10$K & $202$K & No \\
\hline
CASIA-WebFace \cite{CASIA_WebFace}& public & $10$K & $500$K & No \\
\hline
%\bf{Ours} & public & $100$K & about $10000$ K \\
\bf{Ours} & \bf{public} & \bf{$100$K} & about \bf{$10$M} & Yes \\
\hline
Facebook & private & $4$K & $4400$K &Unknown \\
\hline
%Google & private & $8000$ K & $100-200$ M \\
Google & private & $8$M & $100$-$200$M & Unknown\\
\hline
\end{tabular}
\end{center}
\end{table}
\fi

As shown in Table \ref{table:datasets}, our training dataset is considerably
larger than the publicly available datasets. 
Another uniqueness of our training dataset %and other datasets 
is that our dataset focuses on facilitating our celebrity recognition task, 
so our dataset needs to cover as many popular celebrities as possible, 
and have to solve the data disambiguation problem to collect right images for each celebrity. 
On the other hand, the existing datasets are mainly used to train a generalizable face feature, 
and celebrity coverage is not a major concern for these datasets. 
Therefore, for the typical existing dataset, 
if a name string corresponds to multiple celebrities
(e.g., Mike Smith) 
and would lead to ambiguous image search result, 
these celebrities are usually removed from the datasets to help the precision of the collected training data \cite{vgg_face}. 

%\section{Dataset construction}
%\section{One million celebrity list}
\section{Benchmark construction}
\label{sec:dataset}
Our benchmark task is to recognize one million celebrities from their face images, 
and link to their corresponding entity keys in the knowledge base. 
Here we describe how we construct this task in details. 
%This section is to describe the steps to obtain the one-million celebrity list, 
%and steps to construct the measurement and the training set. 

\subsection{One million celebrity list}
We select one million celebrities to recognize
%as our recognition target 
from a knowledge graph called freebase \cite{freebase}, 
where each entity is identified by a unique key (called machine identifier, MID in freebase)
and associated with rich properties. 
We require that the entities we select 
%The celebrities we select need 
%to be 
are
human beings in the real world and have/had public attentions. 

The first step is to select a subset of entities (from freebase \cite{freebase})
which correspond to real people  
using the criteria in \cite{Yandong:Celeb}.  
%Freebase \cite{freebase}
%is chosen since freebase has large coverage, good quality, rich information, and are publicly available. 
In freebase, 
there are 
more than $50$ million topics capsulated in about $2$ billion triplets. 
Note that we don't include any person if his/her facial appearance is 
unknown or not clearly defined. 
\iffalse
%For example, 
This is the reason that 
we do not include movie characters (visual appearance could be subjective)
or the persons 
who are known to be born long time before %1846,
%the time when 
the first roll-film specialized camera ``Kodak'' was invented \cite{Camera} in $1988$.
%in 1888 \cite{Camera}.  
%since we can not rely on drawings or sculptures to recognize people's faces. 
%We can not rely on drawings or sculptures to recognize people's faces, 
%since whether they are visually similar to the actual person could be subjective and arguable. 
%An interesting example is that the sculpture of John Harvard in Harvard university is claimed to be inspired 
%by a Harvard student Sherman Hoar rather than Harvard himself,
%since no one knew what John Harvard had looked like \cite{Harvard}. 
\fi

The second step is to rank 
all the entities in the above subset according to the frequency of their occurrence on the web \cite{Yandong:Celeb}.
%Then, 
We select the top one million entities to form our celebrity list and provide their entity keys (MID) in freebase. 
%The occurrence frequency for a given entity is obtained by counting how many documents contain this entity in a large corpus 
%with billions of documents from the web. 
We concern the public attention (popularity on the web) for two reasons. 
First, 
we want to align our benchmark task with the interest of real applications. 
For %example, 
%in 
applications like image search, image annotations and deep understanding, 
% for news image selection, 
%\leizhangc{do you mean news here?} 
%articles, 
and image caption generation, 
the recognition of popular celebrities would be more attractive to most of the users
than ordinary people. 
Second, we include popular celebrities so that we have better chance to obtain multiple authority images for each of them
to enable our training, testing, and re-distributing under certain licenses. 

\begin{figure}[h]
\centering
%\hspace*{\fill} 
%\subfigure[With sampling weight $f$]{\includegraphics[width=0.90\linewidth]{figs/Histogram2.png}}
%\hspace*{\fill}
\subfigure[Professions]{\includegraphics[width=0.46\linewidth]{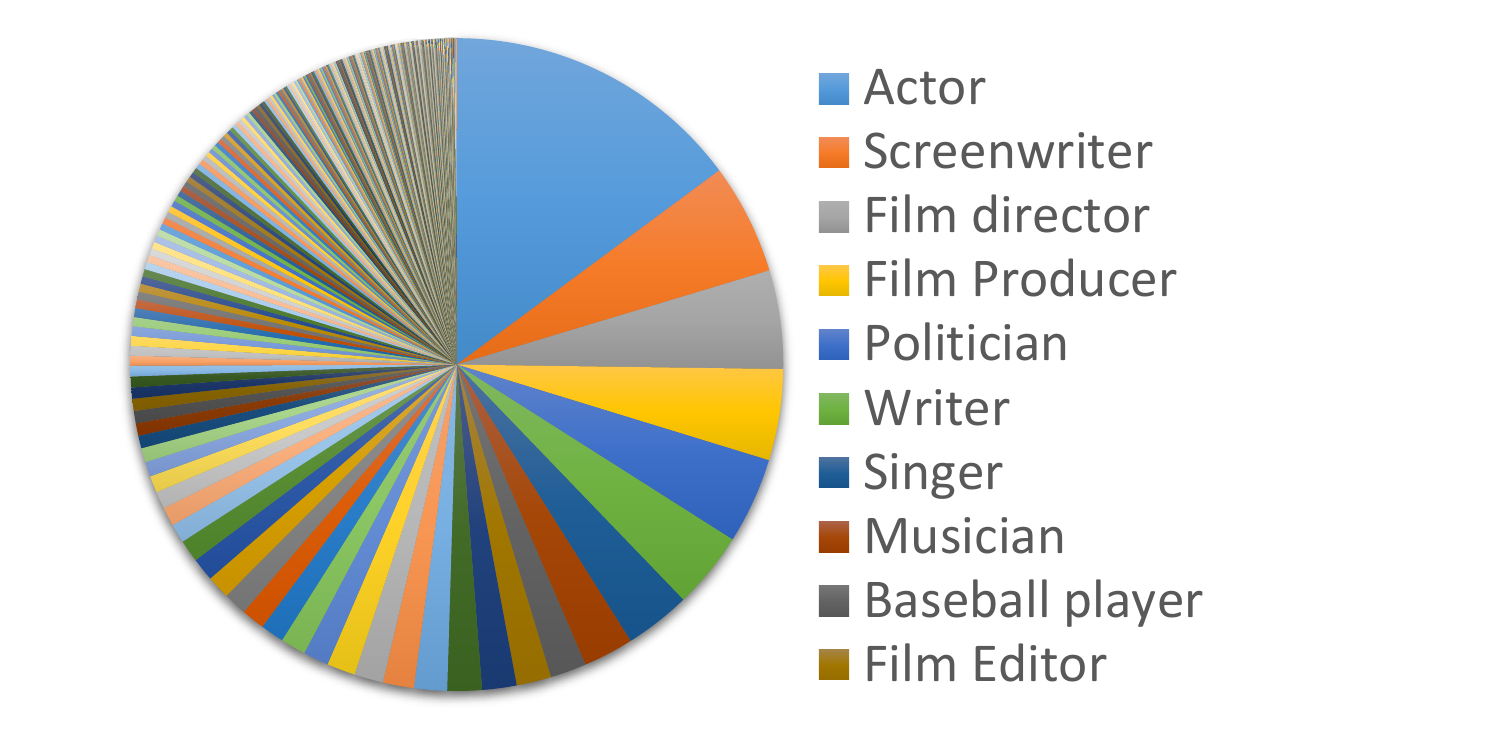}}
\subfigure[Nationality]{\includegraphics[width=0.46\linewidth]{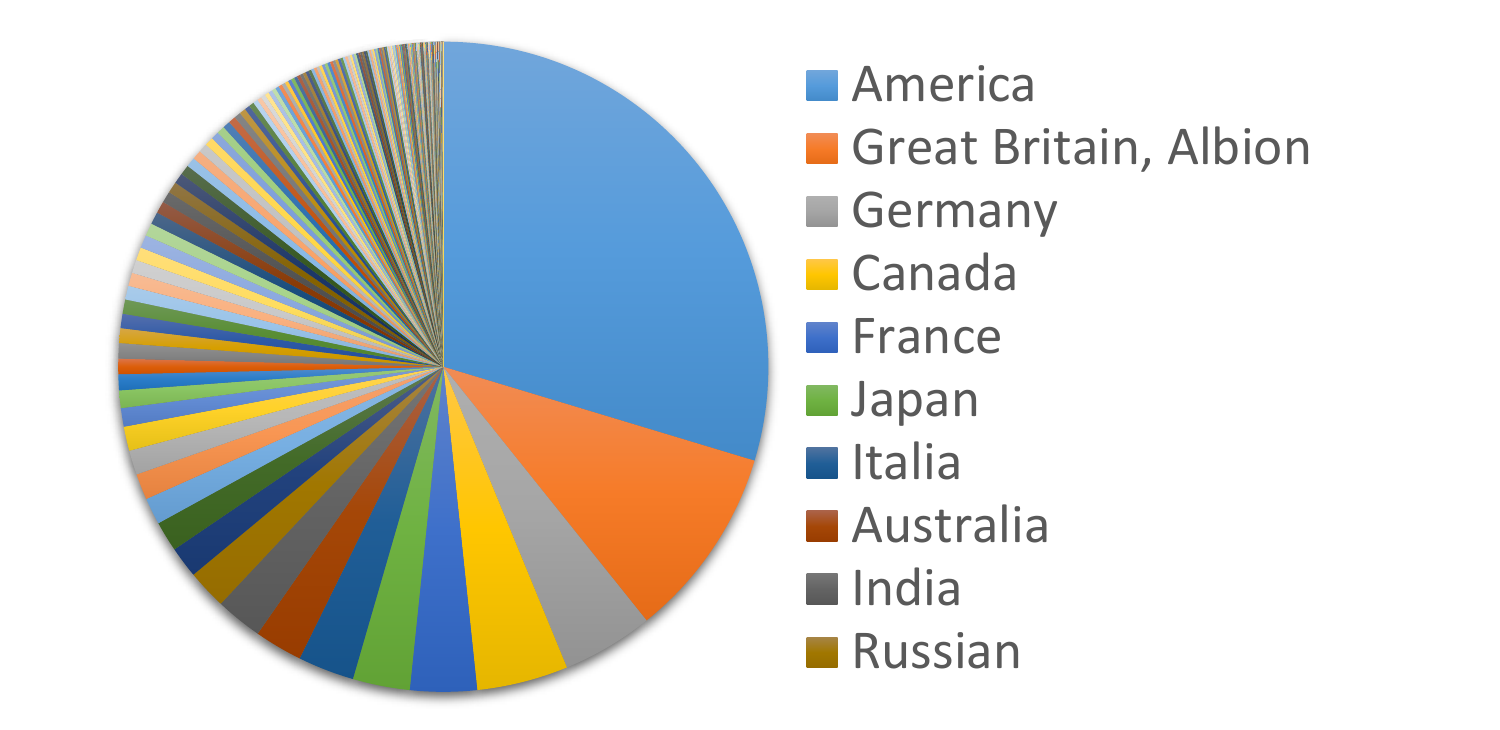}}
%\hspace*{\fill}
%\subfigure[With sampling weight $f'$ defined in (\ref{eq:adjust})]{\includegraphics[width=0.90\linewidth]{figs/Histogram_fp.png}}
\hspace{\fill}
\hspace{\fill}
\subfigure[Date of Birth]{\includegraphics[width=0.60\linewidth]{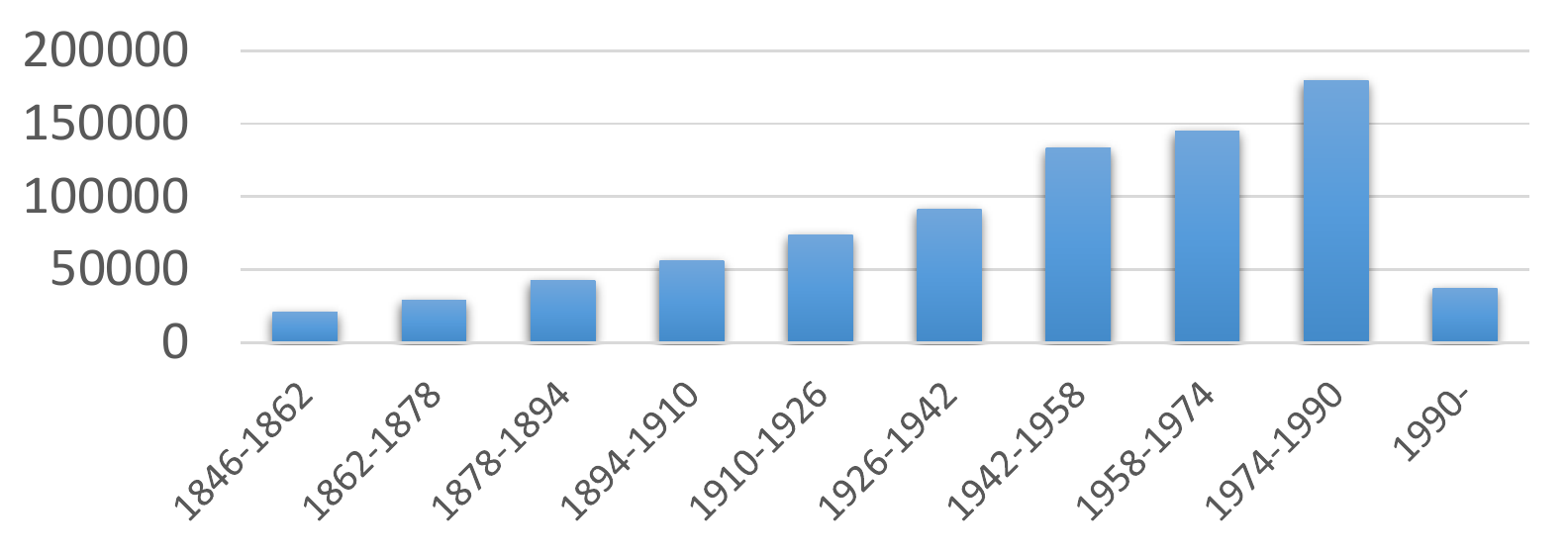}}
\subfigure[Gender]{\includegraphics[width=0.20\linewidth]{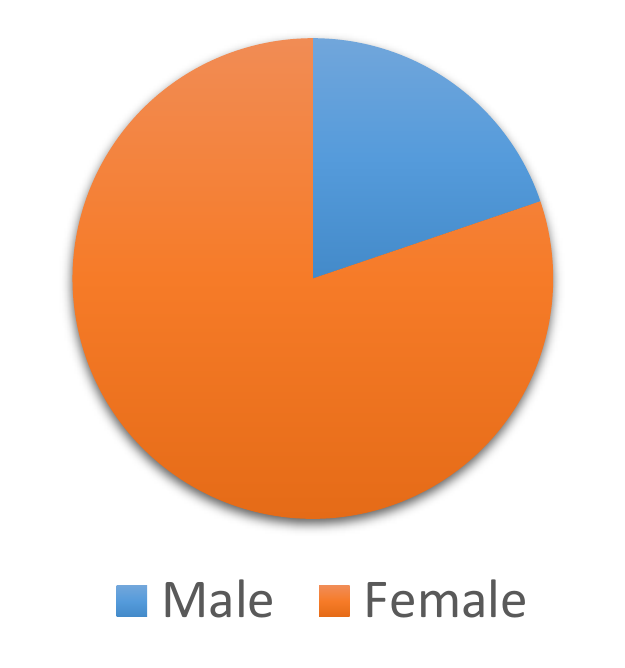}}
%\hspace*{\fill}
\caption{
Distribution of the properties of the celebrities in our one-million list in different aspects. 
The large scale of our dataset naturally introduces great diversity. 
%As shown in (a), the top $10$ professions include actor, screenwriter, film director, film producer, politician, writer, singer, 
%musician, baseball player, and film editor. 
As shown in (a) and (b), we include persons with more than $2000$ different professions, 
and come from more than $200$ distinct countries/regions. 
The figure (c) demonstrates that we don't include celebrities who were born before $1846$ 
(long time before the first roll-film specialized camera ``Kodak'' was invented \cite{Camera})
and covers celebrities of a large variance of age. 
%most of the celebrities in our dataset were born after $1974$. 
In (d), we notice that we have more females than males in our one-million celebrity list.
This might be correlated with the profession distribution in our list. 
}
\label{fig:prop}
\end{figure}

We present the distribution of the one million celebrities in different aspects including profession, nationality, age, and gender. 
In our one million celebrity list, 
we include persons with more than $2000$ different professions (Fig. \ref{fig:prop} (a)), 
and come from more than $200$ distinct countries/regions (Fig. \ref{fig:prop} (b)), 
which introduces a great diversity to our data. 
We cover all the major races in the world (Caucasian, Mongoloid, and Negroid). 
Moreover, 
as shown in Fig. \ref{fig:prop} (c), we cover a large range of ages in our list. 
Though we do not manually select celebrities to make the profession (or gender, nationality, age) distribution uniform, 
the diversity (gender, age, profession, race, nationality) of our celebrity list is guaranteed by the large scale of our dataset. 
This is different from \cite{2015_CVPR_FaceData}, in which there are about $500$ subjects so 
the manual balancing over gender distribution is inevitable. 

Note that our property statistics are limited to the availability of freebase information. 
Some celebrities in our one million list do not have complete properties. 
If 
a certain celebrity does not have property $A$ available in freebase, 
we do not include this celebrity for the statistic calculation of the property $A$. 

\subsection{Celebrity selection for measurement}
In order to evaluate the recognition performance on the one million celebrities obtained in the last subsection, 
we build up a measurement set
which includes 
a set of carefully labeled images
blended with another set of randomly selected face images as distractors. 
The measurement set construction is described in details in the following subsections, 
while 
the evaluation protocol is described in Section \ref{sec:CelebRec}. 

For the labeled images, we 
%applied weighted sampling
%to obtain a subset of celebrities 
sample a subset of celebrities 
\footnote{Currently there are $1500$. 
We will increase the number of celebrities in our measurement set in the future.}
from the one-million celebrity list
due to limited labeling resource. 
The sampling weight is designed in a way that, 
our measurement set mainly focuses on
%head celebrities (celebrities rank top in the occurrence frequency list), 
top celebrities (rank among the top in the occurrence frequency list)
to represent the interest of real applications and users, 
%popularity distribution,
yet maintain a certain amount of tail celebrities (celebrities not mentioned frequently on the web, e.g., from $1$ to $10$ times in total)
to guarantee the measurement coverage over the one-million list. 

More specifically, let $f_i$ denote the number of documents mentioned the $i^{th}$ celebrity on the web. 
Following the method in \cite{Yandong:Celeb}, 
we set 
the probability for the $i^{th}$ celebrity to get selected to be proportional to $f_i'$, defined as, 
\begin{equation}\label{eq:adjust}
f_i' = f_i^{\frac{1}{\sqrt{5}}} \, , 
\end{equation} 
where the exponent $1/\sqrt{5}$ is obtained empirically
to include more celebrities with small $f$. 

Though it seems to be a natural solution, 
we do not set the sampling weights to be proportional to $f_i$, 
since this option will make 
our measurement set barely contain 
any celebrities from the bottom $90 \%$ in our one-million list (ordered by
%web appearance frequency 
$f_i$). 
The reason is that the distribution of 
%the number of appearance of celebrities
$f$
is very long-tailed.
%as shown in Fig. \ref{Figure:f_distribution}. 
More than $90\%$ of the celebrities have $f$ smaller than $30$, while the top celebrities have $f$
larger than one million. 
We need to include sufficient number of tail celebrities to encourage researchers to work on the hard cases to improve the performance
%in terms of recall (or recognition coverage).
from the perspective of recognition coverage. 
This is the reason that we applied the adjustment in (\ref{eq:adjust}). 

%\yandongc{On the other hand, we also need to maintain enough tail celebrities to keep the coverage of our measurement set 
%to encourage researchers to work on the hard cases to improve the performance in terms of recall. }
\iffalse
\begin{figure}
\centering
\includegraphics[width=0.66\linewidth]{figs/LogF_Hist.eps}
\caption{
Distribution of $\log (f)$ (the occurrence frequency of one-million celebrities). 
As shown in the figure, more than $90\%$ of the celebrities have $f$ smaller than $30$, while the top celebrities have $f$
larger than one million. For example, Justin Bieber (entity key m.06w2sn5 in \cite{freebase}) has been mentioned by
$1.9$ millions of distinct documents in our experiment. 
%though still mainly focus on the most popular celebrities. 
}
\label{Figure:f_distribution}
\end{figure}
\fi

%According to our experiments, 
%As reported in \cite{Yandong:Celeb}, 
%\yandongc{polish from here, our recognition task... }
%With the adjustment in (\ref{eq:adjust}), 
%\yandongc{polish from here.}
With the sampling weight $f'$ in 
%Eq.
(\ref{eq:adjust}) applied, 
our measurement set still mainly focuses on the most popular celebrities, 
while about $25\%$ of the celebrities in our measurement set come from the bottom  
%$100$K 
$90 \%$
in our one-million celebrity list (ordered by $f$). 
If we do not apply the adjustment in (\ref{eq:adjust}), but just use $f$ as the sampling weight, 
less than $10\%$ of the celebrities in the measurement set 
come from the bottom $90\%$ in our one-million celebrity list. 
%This would not encourage  researchers 

%Our recognition experiment results empirically demonstrate that the distribution of our measurement set has a good balance 
%in selecting the top/tail entities. 
% \yandongc{well-designed}. 
 
Since the list of the celebrities in our measurement set is not exposed
\footnote{We publish the images for $500$ celebrities, called development set, 
while hold the rest $1000$ for grand challenges.}, 
and our measurement set contains 
$25\%$ of the celebrities in our measurement set come from the bottom  
$90 \%$, 
%Therefore, 
%in order to 
%obtain good coverage performance,
%improve the recognition recall,  
researchers need to include as many celebrities as possible (not only the popular ones) from our one-million list 
to improve the performance
of coverage. 
%to train face recognition model 
This pushes the scale of our task to be very large. 

\subsection{Labeling for measurement}

After we have the set of celebrities for measurement, we provide two images for each of the celebrity.  %by manually labeling. 
%\yandongc{Diverse, no constrained to frontal images.}
The correctness of our image labeling is ensured by
deep research on the web content, %for each of the celebrity, 
multiple iterations of carefully review,
and very rigorous consensus verification. 
%We try our best to select images not in the training dataset to allow our measurement to evaluate the generalization ability of the model to be tested. 
%The image labeling is not constrained by face detection
%since the labeling is completely manual. 
Details are listed as follows. 

\subsubsection{Scraping}
%\noindent
%{\bf{Scraping}}
%The image is from internet scraping. 
Scraping provides image candidates for each of the celebrities selected for the measurement set. 
Though in the end we provide only two images per celebrity for evaluation, we scraped about $30$ images per celebrities. 
During the scraping procedure, 
we applied
different search queries, including the celebrity's name, name plus profession, and names in other languages (if available). 
The advantages of introducing multiple variations of the query used for each celebrity 
is that with multiple queries, we have better chance to capture the images which are truly about the given celebrity. 
Moreover, the variation of the query and scraping multiple images also brings in the diversity to the images for the given celebrity. 
Especially for the famous celebrities, the top one image returned by search engine
is typically his/her representative image (frontal facial image with high quality), which is relatively easier to recognize,
compared with the other images returned by the search engine. 
We increase the scraping depth so that we have more diverse images to be recognized for each of the celebrity. 
%If we always include the top one image, the recognition task can not truly evaluate the ability of the recognition. 

%\iffalse
\begin{figure}
\centering
\includegraphics[width=0.900\linewidth]{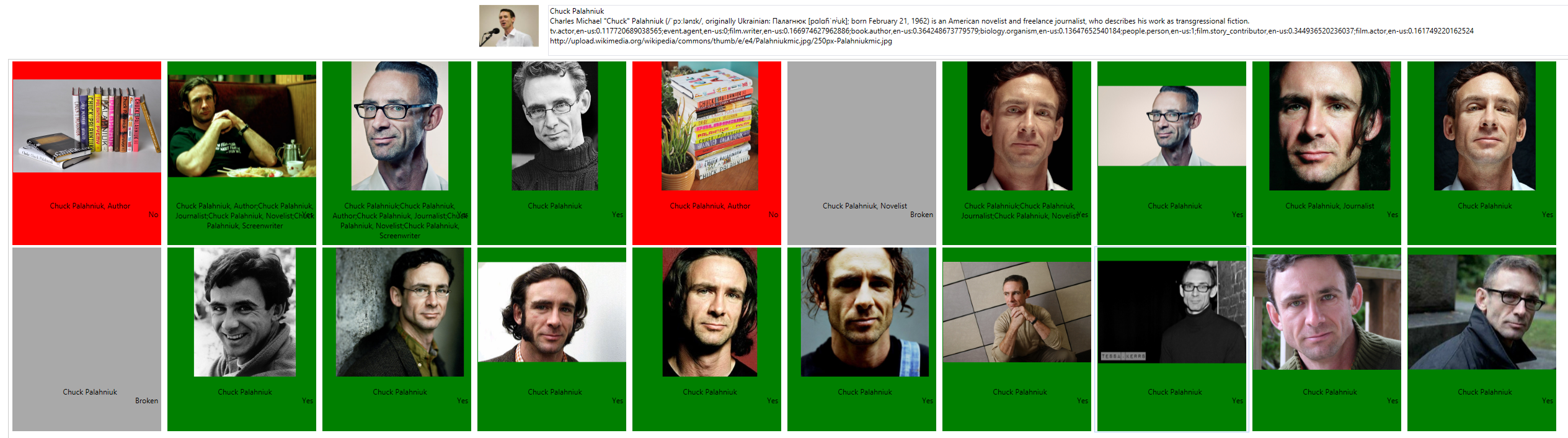}
\caption{
Labeling GUI for ``Chuck Palhniuk''. (partial view)
As shown in the figure, in the upper right corner, 
a representative image and a short description is provided. 
For a given image candidate, judge can label as ``not for this celebrity'' (red), 
``yes for this celebrity'' (green), or ``broken image'' (dark gray). 
%\yandongc{multiple image, multiple views}
%When we label these images,
%a representative image and a short description for the given celebrity is provided for reference,
%as shown in the upper-right corner in Fig.\ref{figure:labeling}. 
%though still mainly focus on the most popular celebrities. 
}
\label{figure:labeling}
\end{figure}
%\fi
\noindent
%\subsubsection{Label}
{\bf{Label}}
Labeling picks up the images which are truly about the given celebrity.
As shown in Fig.\ref{figure:labeling}, 
for each given celebrity, 
we (all the authors)
%\yandongc{shall we use ``judges''} 
manually label {\bf all} the scraped image candidates
%obtained above 
to be truly about this celebrity or not. 
%with 
Extreme cautious was applied.  
We 
have access to the page which contains the scraped image to be labeled. 
Whenever needed, the judge (the authors) is asked to visit the original page with the scraped image and read the page content to guide his/her labeling. 
The rich information on the original page benefits the quality of the labeling, especially for a lot of the hard cases. 
Each of the image-celebrity entity pair was judged by at least two persons. 
Whenever there is a conflict, the two judges review together and provide the final decision based on verbal discussion. 
In total, we have about $30K$ images labeled, spent hundreds of hours. % expert hours. 

In our measurement set, 
we select two images for each of the celebrity to keep the evaluation cost low. 
We have two subset (each of them have the same celebrity list), described as follows. 
\begin{itemize}
\item {\bf{Random set}} 

The image in this subset is randomly selected from the labeled images. 
One image per celebrity. This set reveals how many celebrities are truly covered by the models to be tested. 
\item {\bf{Hard set}}

The image in this subset is the one (from the labeled images) which is the most different from any images in the training dataset. 
One image per celebrity. This set is to
evaluate the generalization ability of the model. 
\end{itemize}
Then, we blend the labeled images with 
%$100$K one
images from other celebrities or ordinary people. 
The evaluation protocol is introduced in details in the next section. 
%The precision and recall are used to evaluate the model performance. 
%We will provide baseline model performance to demonstrate this is a doable task and challenge enough.  
%How to evaluate \yandongc{recall, precision?}. 
%How to rank different teams? 

\section{Celebrity recognition}
\label{sec:CelebRec}

In this section, we set up the evaluation protocol for our benchmark task. %proposed in Section \ref{sec:dataset}. 
Moreover, 
in order to facilitate the researchers to work on this problem, 
we provide a training dataset which is encouraged (optional ) to use. 
We also present 
the baseline performance obtained by using our provided training data. 
We also encourage researchers to train with outside data and evaluate in a separate track. 

\subsection{Evaluation Protocol}
We evaluate the performance of our proposed recognition task in terms of precision and coverage (defined in the following subsection) 
using the settings described as follows. 
%People can use the training data we provided and described in subsection, and/or collect more images as their own training data.  

\subsubsection{Setup}
We setup our evaluation protocol as follows. 
For a model to be tested, 
we collect the model prediction for both the labeled image and distractors in the measurement set. 
%\section \ref{sec:dataset}). 
%Evaluate on both labeled and distractors. 
Note that we don't expose which images in the measurement are labeled ones or which are distractors. 
%Our evaluation protocol has the following setups. 
%First, we don't expose the selected
%$1000$ 
%celebrities in our measurement set. 
%\yandongc{panja here? or mention panja in the GC paper}
%The advantage of not exposing the selected celebrities is that  
This setup avoids human labeling to the measurement set, 
%This setup 
and encourages 
researchers to build a recognizer which could robustly distinguish one million (as many as possible) people faces, 
rather than focusing merely on a small group of people. 

Moreover, 
during the training procedure, if the researcher leverages outside data for training, 
%Second, 
we do not require participants to exclude celebrities in our measurement from the training data set. 
Our measurement still evaluate the generalization ability of our recognition model, 
%the recognition model to have generalization ability, 
due to the following reasons. 
%\yandongc{Requires the  the recognition model to have generalization ability based on the following reasons:}
There are one million celebrities to be recognized in our task, 
and there are millions of images for some popular celebrities on the web. 
It is practically impossible to include all the images for every celebrity in the list. 
On the other hand, according to section \ref{sec:trainingdata}, 
the images in our measurement set is typically not the representative images for the given celebrity (e.g., the top one searching result). 
Therefore the chance to include the measurement images in the training set is relatively low,
as long as the celebrity list in the measurement set is hidden. 
%Second, it is too expensive to manually label all the images for every celebrity in our one-million list. 
This is different from most of the existing face recognition benchmark tasks, 
in which the measurement set is published and targeted on a small group of people. 
For these traditional benchmark tasks, 
%the images of all the persons in the measurement set need to be removed from the training set. 
the evaluation generalization ability relies on manually excluding the images (from the training set) of all the persons in the measurement set
(This is mainly based on the integrity of the participants). 

\subsubsection{Evaluation metric}

In the measurement set, we have $n$ images, denoted by $\{ x_i\}_{i = 1}^n$. 
The first $m$ images $\{ x_i | i = 1, 2, 3, ..., m\}$ are the labeled images for our selected celebrities, 
while the rest $\{ x_i | i = m+1, ..., n\}$ are distractors.
Note that we hide the order of the images in the measurement set. 

For the $i^{th}$ image, let $g(x_i)$ denote the ground truth label (entity key obtained by labeling). For any model to be tested, we assume the model to output $\{ \hat{g}(x_i), c(x_i) \}$ as  
the predicted entity key of the $i^{th}$ image, 
and its corresponding prediction confidence. 
We allow the model to perform rejection. That is, 
if $c(x_i) < t$, where $t$ is a preset threshold, the recognition result for image $x_i$ will be ignored. 
We define the precision with the threshold $t$ as, 
\begin{equation}\label{precision}
P(t) = \frac{|\{ x_i | \hat{g}(x_i) = g(x_i) \wedge c(x_i) \geq t , i = 1, 2, ..., m  \}|}{| \{ x_i | c(x_i) \geq t, i = 1, 2, ..., m\} |} \, ,
\end{equation}
where the nominator %$|\{ x_i | \hat{g}(x_i) = g(x_i) \wedge c(x_i) \leq t , i = 1, 2, ..., m  \}|}{| \{ x_i | c(x_i) \leq t, i = 1, 2, ..., m\}|$ 
is 
%the cardinality of the set 
the number
of the images of which the prediction is correct (and confidence score is larger than the threshold). 
The denominator is the number of images (within the set $\{x_i\}_{i=1}^{m}$) which the model does have prediction (not reject to recognize). 

The coverage in our protocol is defined as
\begin{equation}\label{eq:recall}
C(t) = \frac{|\{ x_i | c(x_i) \geq t, i = 1, 2, ..., m\}|}{m}
\end{equation}

For each given $t$, a pair of precision $P(t)$ and coverage $C(t)$ can be obtained for the model to be tested. 
The precision $P(t)$ is a function of $C(t)$. 
Our major evaluation metric is the maximum of the coverage satisfying the condition of precision, $P(t) \geq P_{min}$.
The value of $P_{min}$ is $0.95$ in our current setup. 
Other metrics and analysis/discussions are also welcomed to report. 
The reason that we prefer a fixed high precision and measure the corresponding coverage is because in many real applications high precision is usually more desirable and of greater value. 

\subsection{Training dataset}
\label{sec:trainingdata}
\begin{figure}[t!]
\centering
  \subfigure[Original Image]{\includegraphics[width=0.90\columnwidth]{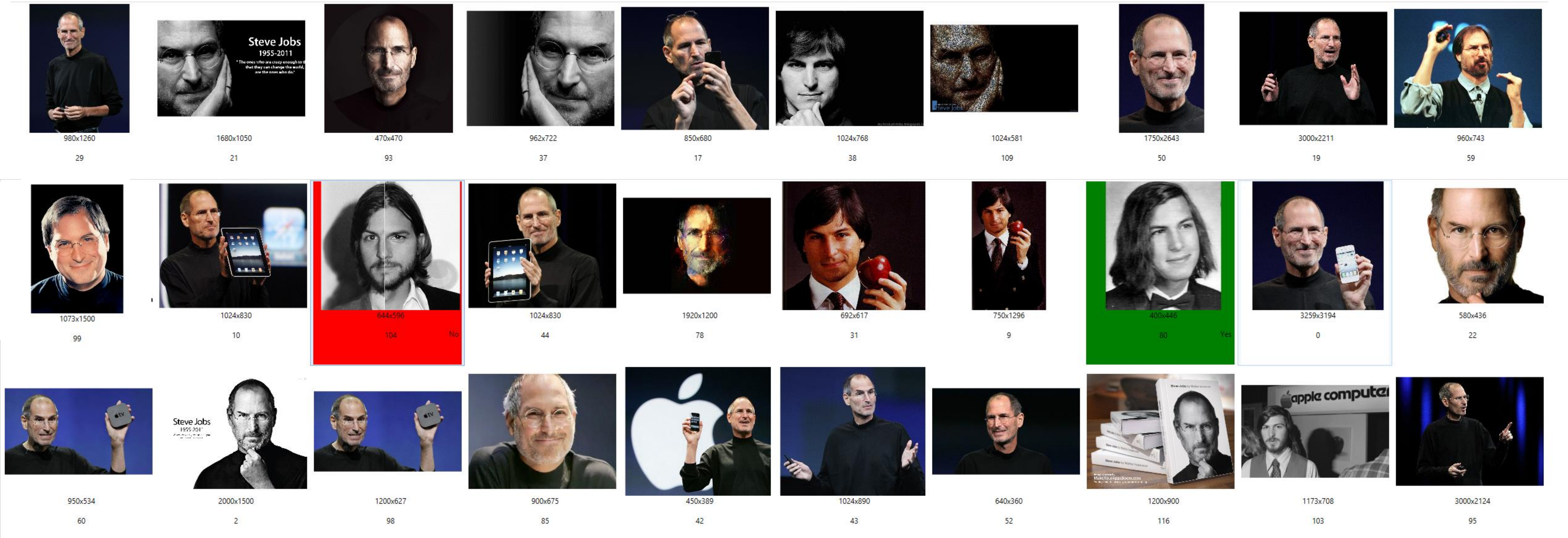}}
  \subfigure[Aligned Face Image]{\includegraphics[width=0.90\columnwidth]{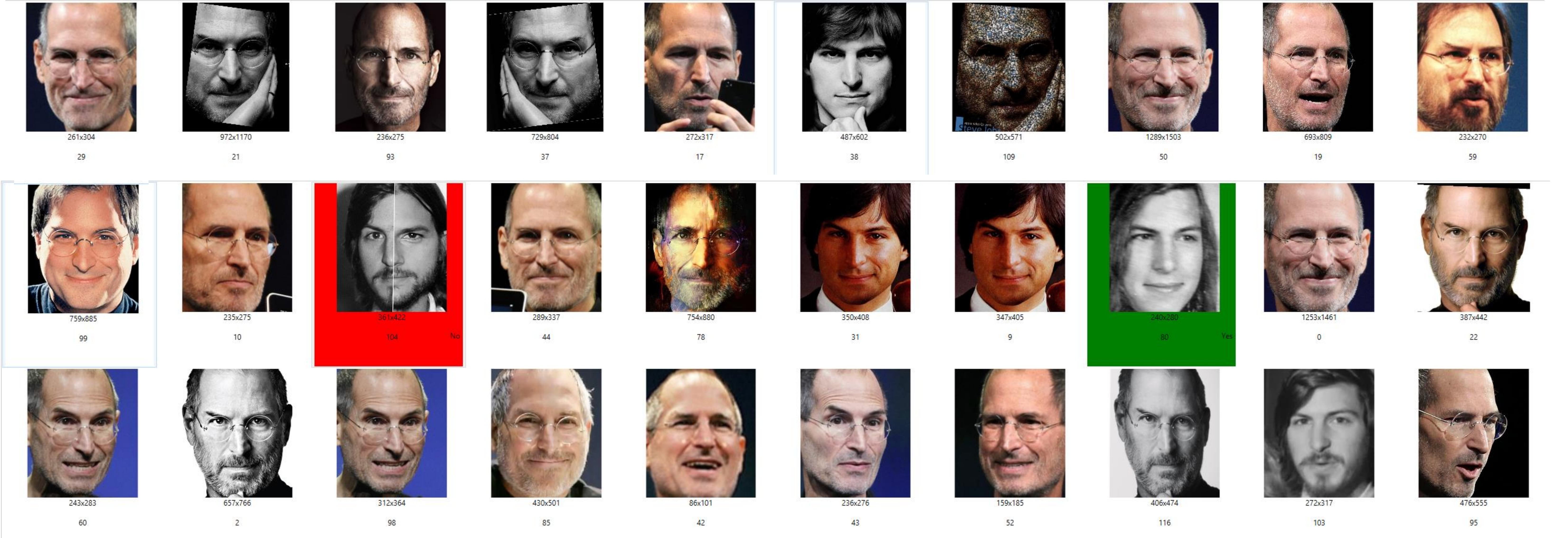}}
  \caption{%Examples of the training images we provided for the celebrity with entity key m.0478\underline{  }\underline{  }m (Lady Gaga). 
Examples (subset) of the training images for the celebrity with entity key m.06y3r (Steve Jobs).
The image marked with a green rectangle is claimed to be Steve Jobs when he was in high school.
The image marked with a red rectangle is considered as a noise sample in our dataset,
since it is synthesized by combining one image of Steve Jobs and one image of Ashton Kutcher, who is the actor in the movie ``Jobs''. }
  \label{Figure:SteveJobs}
\end{figure}

%\yandongc{Mode-based or template-based setup discuss in this section}
In order to facilitate the above face recognition task we provide a large training dataset. 
This training dataset is prepared by the following two steps. 
First, we select the top $100$K entities from our one-million celebrity list in terms of their web appearance frequency. 
Then, we retrieve approximately $100$ images per celebrity from popular search engines. 
%Note these $100$ images are obtained by scraping 
% including Bing and Google, 
%and may contain noises for some of the celebrities. 

%The celebrities in our training data set covers a variaty of professions, covers people borm from 
We do not provide training images for the entire one-million celebrity list for the following considerations. 
First, limited by time and resource, we can only manage to prepare a dataset of top $100$K celebrities as a v1 dataset to facilitate the participants to quickly get started. 
We will continuously extend the dataset to cover more celebrities in the future. 
Moreover, as shown in the experimental results in the next subsection, this dataset is already very promising to use. 
Our training dataset covers about $75\%$ of celebrities in our measurement set, 
which implies that the upper bound of recognition recall rate based on the provided training data cannot exceed $75\%$.
Therefore, we also encourage the participants, especially who are passionate to break this $75\%$ upper bound
to treat the dataset development as one of the key problems in this challenge, 
and bring in outside data to get higher recognition recall rate and compare experimental results in a separate track. 
Especially, we encourage people label their data with entity keys in the freebase snapshot we provided and publish, so that different dataset could be easily united to facilitate collaboration. 

On example in our training dataset is shown in Figure \ref{Figure:SteveJobs}. 
As shown in the figures, %\ref{Figure:LadyGaga}, 
same celebrity may look very differently in different images. 
%from image to image. 
In Figure \ref{Figure:SteveJobs}, we see images for Steve Jobs (m.06y3r) when he was about $20/30$ years old, as well as images when he was about $50$ years old.
The image at row 2, column 8 (in green rectangle) in Figure \ref{Figure:SteveJobs} is claimed to be Steve Jobs when he was in high school. 
%We see a good age span in our dataset. 
Notice that the image at row 2, column 3 in Figure \ref{Figure:SteveJobs}, marked with red rectangle is considered as a noise sample in our dataset,
since this image was synthesized by combining one image of Steve Jobs and one image of Ashton Kutcher,
who is the actor in the movie ``Jobs''. 

As we have mentioned, we do not manually remove the noise in this training data set. 
This is partially because to prepare training data of this size is beyond the scale of manually labeling. 
In addition, we have observed that the state-of-the-art deep neural network learning algorithm can tolerate a certain level of noise in the training data. Though for a small percentage of celebrities their image search result is far from perfect, 
more data especially more individuals covered by the training data could still be of great value to the face recognition research, 
which is also reported in \cite{vgg_face}. 
Moreover, 
we believe that data cleaning, noisy label removal, and learning with noisy data are all good and real problems that are worth of dedicated research efforts. 
Therefore, we leave this problem open and do not limit the use of outside training data.

\subsection{Baseline}

There are typically two categories of methods to recognize people from face images. 
One is template-based. For methods in this category, a gallery set which contains multiple images for the targeted group of people is pre-built. 
Then, for the given image in the query set, the most similar image(s) in the gallery set (according to some certain metrics or in pre-learned feature space) is retrieved, and the annotation of this/these similar images are used to estimate the identity of the given query image. 
When the gallery is not very large, 
this category of methods is very convenient for adding/removing entities in the gallery since the face feature representation could be learned in advance. 
However, when the gallery is large, a complicated index needs to be built to shorten the retrieval time. 
In this case, the flexibility of adding/removing entities for the methods in this category vanishes. 
Moreover, the accuracy of the template-based methods highly relies on the annotation accuracy in the gallery set. 
When there are many people in the targeted group, accurate annotation is beyond human effort and could be a very challenging problem itself. 

We choose the second category, which is a model-based method. 
More specifically, we model our problem as a classification problem 
and consider each celebrity as a class. % in the classification problem. 

\begin{figure}
\centering
\subfigure[Original Curve]{\includegraphics[width=0.36\columnwidth]{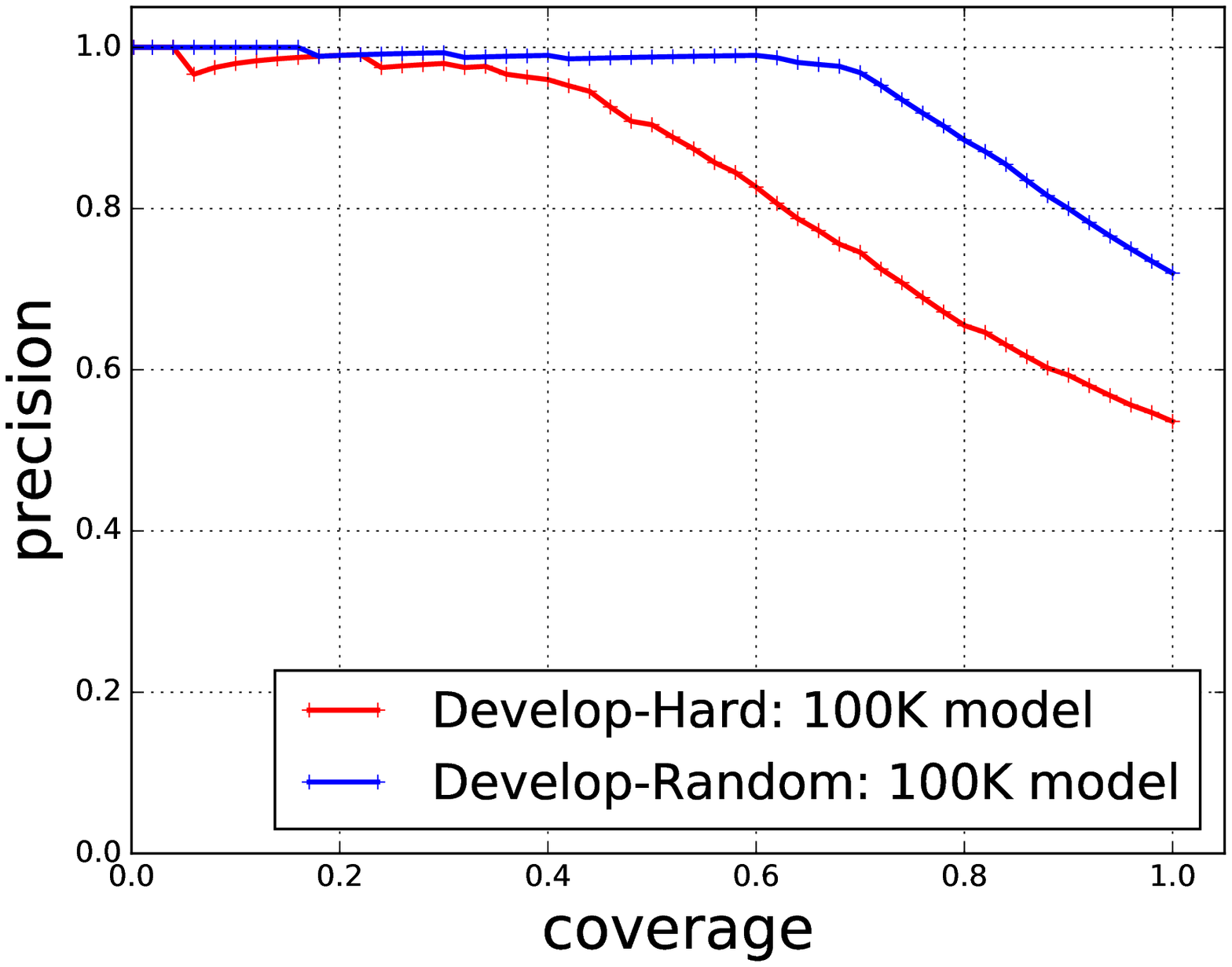}}
\subfigure[Zoom-in]{\includegraphics[width=0.36\columnwidth]{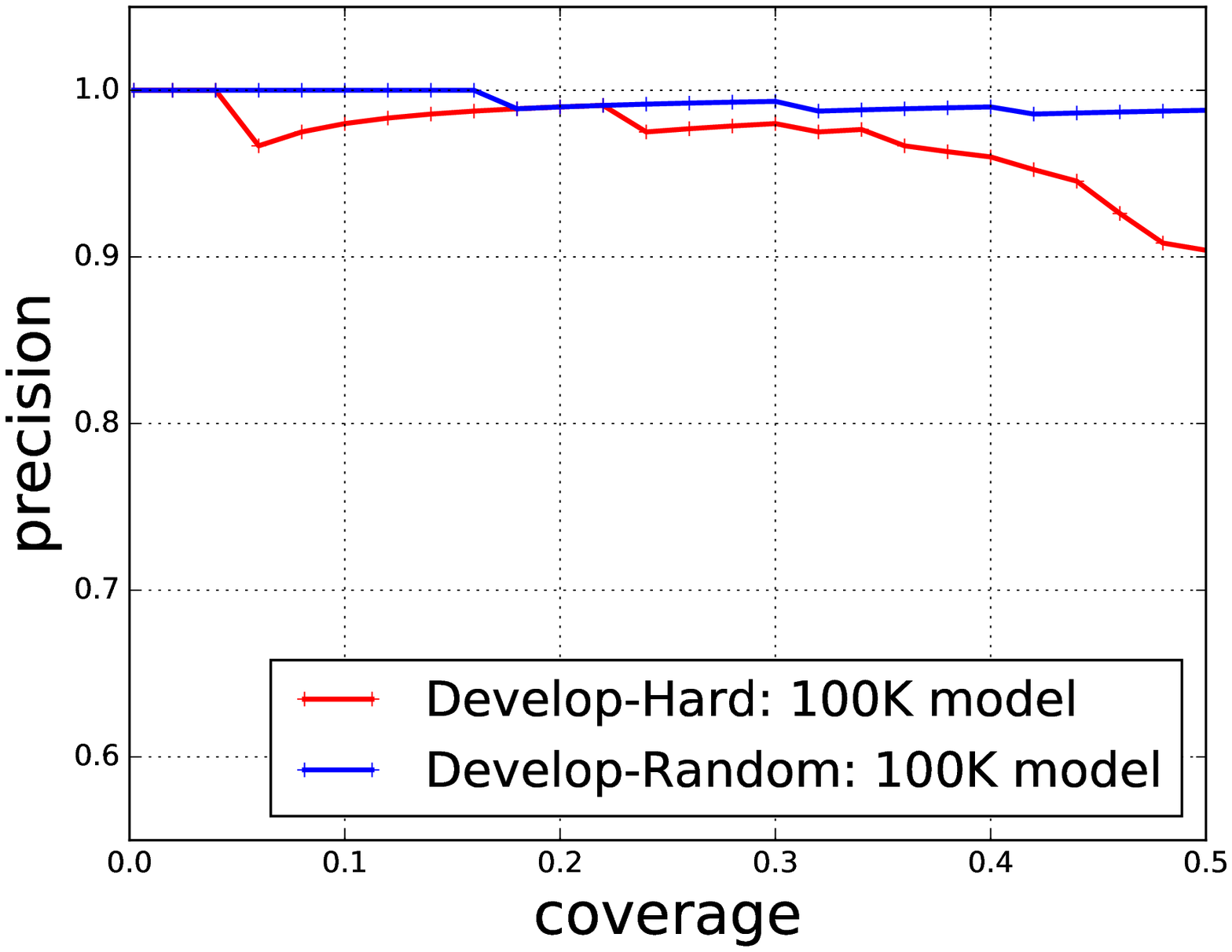}}
  \caption{Precision-coverage curve with our baseline model}
\label{fig:pcCurve}
\end{figure}

In our experiment, we 
%applied simple clean up to the training data \yandongc{no clean up?}
% as described in \ref{sec:trainingdata} and 
trained a deep neural network following the network structure in \cite{AlexHinto_DNN}. 
Training a deep neural network for $100$K celebrities is not a trivial task. 
If we directly train the model from scratch, it is hard to see the model starts to converge even after a long run due to the large number of categories. 
To address this problem, we started from training a small model for $500$ celebrities, which have the largest numbers of images for each celebrity. 
In addition, we used the pre-trained model from \cite{AlexHinto_DNN} to initialize this small model. This step is optional, but we observed that it helps the training process converge faster. After $50,000$ iterations, we stopped to train this model, and used it as a pre-trained model to initialize the full model of $100$K celebrities. 
After $250,000$ iterations, 
with learning rate decreased from the initial value $0.01$ to $0.001$ and $0.0001$ after $100,000$ and $200,000$ iterations, the training loss decrease becomes very slow and indiscernible. 
Then we stopped the training and used the last model snapshot to evaluate the performance of celebrity recognition on our measurement set.
%We applied simple clean up to the training data described in \ref{sec:training data} as follows and trained a deep neural network following the network structure in \cite{AlexHinto_DNN}. 
The experimental results (on the published $500$ celebrities) are shown in Fig. \ref{fig:pcCurve} and Table \ref{table:pcCurve}.

%As shown in the figure, 
%with the \textit{hard} subset, 
%the coverage of this baseline method corresponding to precision $95\%$ is $0.44$, 
%while with the \textit{random} subset, 
%the coverage of this baseline method corresponding to precision $95\%$ is $0.73$.  

%This simple baseline performance is pretty promising, as we can see the model can achieve a very high precision of for a decent coverage (For \textit{random} subset, even with precision $99\%$, the coverage is $0.60$.). 
\begin{table}
\caption{Experimental results on the $500$ published celebrities}
\label{table:pcCurve}
%\centering
\begin{center} 
%% Some packages, such as MDW tools, offer better commands for making tables
%% than the plain LaTeX2e tabular which is used here.
\begin{tabular}{|c||c|c|}
\hline
 & Coverage@Precision $99\%$ &  Coverage@Precision $95\%$  \\
\hline
%\bf{Ours} & public & $100$K & about $10000$ K \\
\textit{Hard} Set & $0.052$ & $0.442$   \\
\hline
\textit{Random} Set & $0.606$ & $0.728$  \\
\hline
\end{tabular}
\end{center}
\end{table}

The promising results can be attributed to the deep neural network capability 
and the high quality of image search results thanks for years of improvement in image search engines. 
However, the curves also shows that the task is indeed very challenge. To achieve both high precision and high recall, a great amount of research efforts need to be spent on data collection, cleaning, learning algorithm, and model generalization, which are valuable problems to computer vision researchers.

%while the coverage of our method corresponding to FPIR $1/100$ is $y$. 

\section{Discussion and Future work}

In this paper, we have defined
a benchmark task which is 
to recognize one million celebrities in the world from their face images, 
and link the face to a corresponding entity key in a knowledge base. 
Our face recognition has the property of disambiguation, and close to the human behavior in recognizing images. 
We also provide concrete measurement set for people to evaluate the model performance easily, 
and provide, to the best of our knowledge, the largest training dataset to facilitate research in the area. 

Beyond face recognition, our datasets could inspire other research topics. 
For example, people could adopt one of the cutting-edge unsupervised/semi-supervised clustering algorithms 
%\cite{Ng01onspectral, Belkin:SSL, Zhu-SSL, Zhou04learningwith}
\cite{Ng01onspectral}
\cite{Belkin:SSL}
\cite{Zhu-SSL}
\cite{Zhou04learningwith}
on our training dataset, 
and/or develop new algorithms which can accurately locate and remove 
outliers in a large, real dataset. 
%The dataset we prepared could inspire research beyond face recognition. 
%For example, how to automatically clean up this training data set is good chance to test any supervised/semisupervised clustering algorithm 
%in a large scale environment. 
Another interesting topic is the to build estimators to predict a person's properties from his/her face images. 
For example, the images in our training dataset are associated with entity keys in knowledge base, of which the gender information (or other properties)
could be easily retrieved. 
People could train a robust gender classifier for the face images in the wild based on this large scale training data. 
We look forward to exciting research inspired by our training dataset and benchmark task. 
\clearpage

\bibliographystyle{splncs}
\bibliography{face}

\begin{thebibliography}{10}

\bibitem{Yandong:Celeb}
Guo, Y., Zhang, L., Hu, Y., He, X., Gao, J.:
\newblock M{S}-{C}eleb-1{M}: Challenge of recognizing one million celebrities
  in the real world.
\newblock In: IS\&T International Symposium on Electronic Imaging. (2016)

\bibitem{FaceBook_2014}
Taigman, Y., Yang, M., Ranzato, M., Wolf, L.:
\newblock Deepface: Closing the gap to human-level performance in face
  verification.
\newblock In: Proc. of IEEE Computer Soc. Conf. on Computer Vision and Pattern
  Recognition (CVPR). (June 2014)

\bibitem{FaceBook_2015}
Taigman, Y., Yang, M., Ranzato, M., Wolf, L.:
\newblock Web-scale training for face identification.
\newblock In: Proc. of IEEE Computer Soc. Conf. on Computer Vision and Pattern
  Recognition (CVPR), IEEE (2015)  2746--2754

\bibitem{Google_Face}
Schroff, F., Kalenichenko, D., Philbin, J.:
\newblock Facenet: A unified embedding for face recognition and clustering.
\newblock In: Proc. of IEEE Computer Soc. Conf. on Computer Vision and Pattern
  Recognition (CVPR). (June 2015)

\bibitem{freebase}
Google:
\newblock Freebase data dumps.
\newblock \url{https://developers.google.com/freebase/data} (2015)

\bibitem{ILSVRC15}
Russakovsky, O., Deng, J., Su, H., Krause, J., Satheesh, S., Ma, S., Huang, Z.,
  Karpathy, A., Khosla, A., Bernstein, M., Berg, A.C., Fei-Fei, L.:
\newblock {ImageNet Large Scale Visual Recognition Challenge}.
\newblock International Journal of Computer Vision (IJCV) \textbf{115}(3)
  (2015)  211--252

\bibitem{LFWTech}
Huang, G.B., Ramesh, M., Berg, T., Learned-Miller, E.:
\newblock Labeled faces in the wild: A database for studying face recognition
  in unconstrained environments.
\newblock Technical Report 07-49, University of Massachusetts, Amherst (October
  2007)

\bibitem{LFWTechUpdate}
Huang, G.B., Learned-Miller, E.:
\newblock Labeled faces in the wild: Updates and new reporting procedures.
\newblock Technical Report UM-CS-2014-003, University of Massachusetts, Amherst
  (May 2014)

\bibitem{Xiaoou_Deep3}
Sun, Y., Wang, X., Tang, X.:
\newblock Deep{I}{D}3: Face recognition with very deep neural networks.
\newblock arXiv preprint arXiv:1502.00873 (2014)

\bibitem{FacePP_ACM}
Fan, H., Yang, M., Cao, Z., Jiang, Y., Yin, Q.:
\newblock Learning compact face representation: Packing a face into an int32.
\newblock In: Proc. of ACM Int'l Conf. on Multimedia, ACM (2014)  933--936

\bibitem{UW_MegaFace}
{Kemelmacher-Shlizerman}, I., {Seitz}, S., {Miller}, D., {Brossard}, E.:
\newblock The {Me}ga{F}ace benchmark: 1 million faces for recognition at scale.
\newblock ArXiv e-prints (2015)

\bibitem{FaceScrub}
Ng, H.W., Winkler, S.:
\newblock A data-driven approach to cleaning large face datasets.
\newblock In: Proc. of IEEE Int'l Conf. on Image Proc. (ICIP). (Oct 2014)

\bibitem{FGNet}
Panis, G., Lanitis, A.:
\newblock An overview of research activities in facial age estimation using the
  {FG-NET} aging database.
\newblock In: Proc. of the European Conf. on Computer Vision (ECCV) Workshops.
  (2014)

\bibitem{Youtube}
Wolf, L., Hassner, T., Maoz, I.:
\newblock Face recognition in unconstrained videos with matched background
  similarity.
\newblock In: Proc. of IEEE Computer Soc. Conf. on Computer Vision and Pattern
  Recognition (CVPR). (2011)

\bibitem{Xiaoou_Deep1}
Sun, Y., Wang, X., Tang, X.:
\newblock Deep learning face representation from predicting 10,000 classes.
\newblock In: Proc. of IEEE Computer Soc. Conf. on Computer Vision and Pattern
  Recognition (CVPR). (June 2014)

\bibitem{CASIA_WebFace}
Yi, D., Lei, Z., Liao, S., Li, S.Z.:
\newblock Learning face representation from scratch.
\newblock arXiv preprint arXiv:1411.7923 (2014)

\bibitem{2015_CVPR_FaceData}
Klare, B.F., Klein, B., Taborsky, E., Blanton, A., Cheney, J., Allen, K.,
  Grother, P., Mah, A., Jain, A.K.:
\newblock Pushing the frontiers of unconstrained face detection and
  recognition: Iarpa janus benchmark a.
\newblock In: Proc. of IEEE Computer Soc. Conf. on Computer Vision and Pattern
  Recognition (CVPR). (June 2015)

\bibitem{vgg_face}
Parkhi, O.M., Vedaldi, A., Zisserman, A.:
\newblock Deep face recognition.
\newblock In: Proceedings of the British Machine Vision Conference (BMVC).
  (2015)

\bibitem{Camera}
Eastman, G.:
\newblock Camera.
\newblock US Patent 388850 A (1888)

\bibitem{AlexHinto_DNN}
Krizhevsky, A., Sutskever, I., Hinton, G.E.:
\newblock Imagenet classification with deep convolutional neural networks.
\newblock In: Advances in Neural Information Processing Systems (NIPS), MIT
  Press (2012)  1097--1105

\bibitem{Ng01onspectral}
Ng, A.Y., Jordan, M.I., Weiss, Y.:
\newblock On spectral clustering: Analysis and an algorithm.
\newblock In: Advances in Neural Information Processing Systems (NIPS), MIT
  Press (2001)  849--856

\bibitem{Belkin:SSL}
Belkin, M., Niyogi, P.:
\newblock Semi-supervised learning on riemannian manifolds.
\newblock Journal of Machine Learning \textbf{56}(1-3) (June 2004)  209--239

\bibitem{Zhu-SSL}
Zhu, X., Ghahramani, Z., Lafferty, J.:
\newblock Semi-supervised learning using gaussian fields and harmonic
  functions.
\newblock In: Proc. of Int'l Conf. on Machine Learning. (2003)  912--919

\bibitem{Zhou04learningwith}
Zhou, D., Bousquet, O., Lal, T.N., Weston, J., Schölkopf, B.:
\newblock Learning with local and global consistency.
\newblock In: Advances in Neural Information Processing Systems (NIPS), MIT
  Press (2004)  321--328

\end{thebibliography}
\end{document}